\documentclass[10pt,twocolumn,letterpaper]{article}

\usepackage{iccv}
\usepackage{times}
\usepackage{epsfig}
\usepackage{graphicx}
\usepackage{amsmath}
\usepackage{amssymb}

\usepackage{multirow}
\usepackage{enumitem}

\usepackage{soul}

\usepackage[pagebackref=true,breaklinks=true,letterpaper=true,colorlinks,bookmarks=false]{hyperref}
\expandafter\def\expandafter\UrlBreaks\expandafter{\UrlBreaks
  \do\a\do\b\do\c\do\d\do\e\do\f\do\g\do\h\do\i\do\j%
  \do\k\do\l\do\m\do\n\do\o\do\p\do\q\do\r\do\s\do\t%
  \do\u\do\v\do\w\do\x\do\y\do\z\do\A\do\B\do\C\do\D%
  \do\E\do\F\do\G\do\H\do\I\do\J\do\K\do\L\do\M\do\N%
  \do\O\do\P\do\Q\do\R\do\S\do\T\do\U\do\V\do\W\do\X%
  \do\Y\do\Z}

\iccvfinalcopy 

\def\iccvPaperID{****} 

\ificcvfinal\fi

\newenvironment{packed_itemize}{
\vspace{-0.15cm}\begin{itemize}
  \setlength{\itemsep}{1pt}
  \setlength{\parskip}{0pt}
  \setlength{\parsep}{0pt}
}{\end{itemize}}

\newenvironment{packed_enumerate}{
\begin{enumerate}
  \setlength{\itemsep}{1pt}
  \setlength{\parskip}{0pt}
  \setlength{\parsep}{0pt}
}{\end{enumerate}}

\begin{document}

\title{Robust Multi-Modality Multi-Object Tracking}

\author{Wenwei Zhang$^1$, Hui Zhou$^{2}$, Shuyang Sun$^{3}$, Zhe Wang$^{2}$, Jianping Shi$^{2}$, Chen Change Loy$^{1}$ \\
  $^{1}$Nanyang Technological University, $^{2}$SenseTime Research, $^{3}$University of Oxford\\
  {\tt\small $\left \{\text{wenwei001, ccloy} \right\}$@ntu.edu.sg, 
  $\left \{\text{zhouhui, wangzhe, shijianping}\right\}$@sensetime.com} \\
  {\tt\small \text{shuyang.sun}@eng.ox.ac.uk}
}

\maketitle
\ificcvfinal\thispagestyle{empty}\fi
	

\begin{abstract}
Multi-sensor perception is crucial to ensure the reliability and accuracy in autonomous driving system, while multi-object tracking (MOT) improves that by tracing sequential movement of dynamic objects. Most current approaches for multi-sensor multi-object tracking are either lack of reliability by tightly relying on a single input source (e.g., center camera), or not accurate enough by fusing the results from multiple sensors in post processing without fully exploiting the inherent information. 
In this study, we design a generic sensor-agnostic multi-modality MOT framework (mmMOT), where each modality (i.e., sensors) is capable of performing its role independently to preserve reliability, and further improving its accuracy through a novel multi-modality fusion module.
Our mmMOT can be trained in an end-to-end manner, enables joint optimization for the base feature extractor of each modality and an adjacency estimator for cross modality. 
Our mmMOT also makes the first attempt to encode deep representation of point cloud in data association process in MOT. 
We conduct extensive experiments to evaluate the effectiveness of the proposed framework on the challenging KITTI benchmark and report state-of-the-art performance. 
Code and models are available at \url{https://github.com/ZwwWayne/mmMOT}.
\end{abstract}
  
\section{Introduction}

\begin{figure}
 \begin{center}
\includegraphics[width=\linewidth]{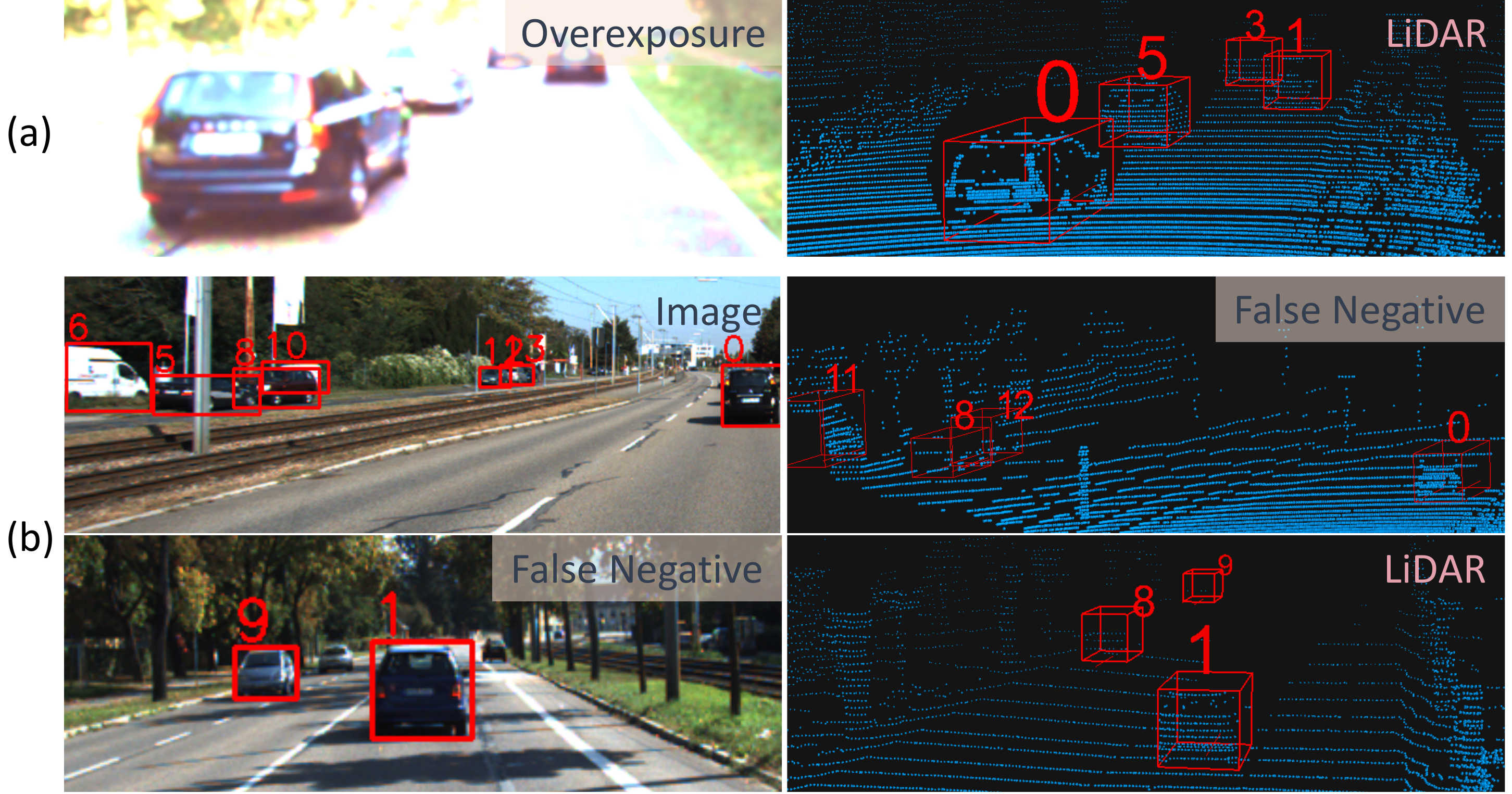}\\
 \vspace{-6pt}
 \end{center}
    \caption{Figure (a) For reliability: camera is disabled when overexposure or crashed in transmission. Figure (b) For accuracy: multi-sensor information could reinforce the perception ability. The image is cropped and best viewed in color and zoomed in.}
 \label{fig:comparison_1}
 \vspace{-12pt}
 \end{figure}

Reliability and accuracy are the two fundamental requirements for autonomous driving system. 
Dynamic object perception is vital for autonomous driving. To improve its reliability, multi-modality sensors can be employed to provide loosely coupled independent clues to prevent failure showed in Figure \ref{fig:comparison_1} (a). To improve accuracy, sequential information from multiple object tracking can be incorporated, and better multi-sensor information can reinforce the final score as in Figure \ref{fig:comparison_1} (b). In this paper, we propose the \textit{multi-modality Multi-Object Tracking}~(mmMOT) framework, which preserves reliability by a novel fusion module for multiple sensors and improves accuracy with attention guided multi-modality fusion mechanism.

It is non-trivial for traditional methods to design a multi-modality (i.e., multi-sensor) MOT framework and preserve both reliability and accuracy. 
A majority of traditional methods \cite{Asvadi20163DOT, ChoSKR14, GarciaA16, osep2017combined} use camera, LiDAR or radar with hand-crafted features fused by Kalman Filter or Bayesian framework. Their accuracy is bounded by the expression ability of hand-crafted features. 
Another stream of methods uses deep feature extractors~\cite{FrossardU18}, which significantly improve the accuracy. Nevertheless, they focus on image level deep representation to associate object trajectories and use LiDAR only in detection stage. Such a binding method cannot preserve reliability if the camera is down.

In this work, we design a multi-modality MOT (mmMOT) framework that is extendable to camera, LiDAR and radar. Firstly, it obeys a loose coupling scheme to allow high reliability during the extraction and fusion of multi-sensor information. Specifically, multi-modality features are extracted from each sensor independently, then a fusion module is applied to fuse these features, and pass them to an adjacency estimator, which is capable of performing inference based on each modality. 
Second, to enable the network to learn to infer from different modalities simultaneously, our mmMOT is trained in an end-to-end manner, so that the multi-modality feature extractor and cross-modality adjacency estimator are jointly optimized.   
Last but not least, we make the first attempt of using deep representation of point cloud in the data association process for MOT and achieve competitive results.

We conduct extensive experiments on the fusion module and evaluate our framework on the  KITTI tracking dataset \cite{Geiger2012CVPR}. Without bells and whistles, we achieve state-of-the-art results on KITTI tracking benchmark \cite{Geiger2012CVPR} under the online setting, purely relying on image and point cloud, and our results with single modality (under sensor failure condition) by the same model are also competitive (only 0.28\% worse). 

To summarize, our contributions are as follows: 
\begin{packed_enumerate}
\item We propose a multi-modality MOT framework with a robust fusion module that exploits multi-modality information to improve both reliability and accuracy.
\item We propose a novel end-to-end training method that enables joint optimization of cross-modality inference.
\item We make the first attempt to apply deep features of point cloud for tracking and obtain competitive results. 
\end{packed_enumerate}

\section{Related Work}
\noindent\textbf{Multi-Object Tracking Framework.}
Recent research of MOT primarily follows the tracking-by-detection paradigm \cite{BreitensteinRLKG11, FrossardU18, BeyondPixels_ICRA2018, zhou_dccrf}, where object of interests is first obtained by an object detector and then linked into trajectories via data association. The data association problem could be tackled from various perspectives, e.g., min-cost flow \cite{FrossardU18, Lenz_2015, Schulter_2017}, Markov decision processes (MDP) \cite{XiangAS15}, partial filtering \cite{BreitensteinRLKG11}, Hungarian assignment \cite{BeyondPixels_ICRA2018} and graph cut \cite{TangAAS15, ZamirDS12}. 
However, most of these methods are not trained in an end-to-end manner thus many parameters are heuristic (e.g., weights of costs) and susceptible to local optima. 

To achieve end-to-end learning within the min-cost flow framework, Schulter et al. \cite{Schulter_2017} applies bi-level optimization by smoothing the linear programming and Deep Structured Model (DSM) \cite{FrossardU18} exploits the hinge loss. Their frameworks, however, are not designed for cross-modality. We solve this problem by adjacency matrix learning. 

%
Apart from different data association paradigms, correlation features have also been explored widely to determine the relation of detections.
Current image-centric methods \cite{FrossardU18, Sadeghian_2017, BeyondPixels_ICRA2018, zhou_dccrf} mainly use deep features of image patches.
Hand-crafted features are occasionally used as auxiliary inputs, including but not limited to bounding box \cite{GunduzA18}, geometric information \cite{PosseggerMRB14}, shape information \cite{BeyondPixels_ICRA2018} and temporal information \cite{tian2019online}. 
3D information is also beneficial and thus exploited by prediction from 3D detection \cite{FrossardU18} or estimation from RGB image with either neural networks \cite{Scheidegger_2018} or geometric prior \cite{BeyondPixels_ICRA2018}. 
Osep et al. \cite{osep2017combined} fuses the information from RGB images, stereo, visual odometry, and optionally scene flow, but it cannot be trained in an end-to-end manner.
All the aforementioned methods must work with camera thus lack of reliability.
By contrast, our mmMOT extracts feature from each sensor independently (both deep image features and deep representation of point cloud), and each sensor plays an equally important role and they can be decoupled. The proposed attention guided fusion mechanism further improves accuracy.

\noindent\textbf{Deep Representation of Point Cloud.}
A traditional usage of point cloud for tracking is to measure distances~\cite{blind_spots}, provide 2.5D grid representation \cite{AsvadiPN15, ChoiULM13} or to derive some hand-crafted features \cite{song2015object}. None of them fully exploit the inherent information of the point cloud for the data association problem. 
Recent studies~\cite{BaiMHWLU18, pmlr-v87-casas18a, LuoYU18} have demonstrated the value of using 3D point cloud as perception features in autonomous driving. 
To learn a good deep representation for point cloud, PointNet \cite{Charles_2017} and PointNet++ \cite{qi2017pointnet} process raw unstructured point clouds using symmetric functions. We adopt this effective method in our framework. 
Other studies such as PointSIFT \cite{PointSIFT} proposes an orientation-encoding unit to learn SIFT-like features of point cloud, and 3DSmoothNet \cite{3DSmoothNet} learns a voxelized smoothed density value representation. 
There are also methods \cite{Wu_2018, SqueezeSegV2} which project the point cloud to a sphere thus 2D CNN can be applied for the segmentation task.

\noindent\textbf{Object Detection.}
An object detector is also a vital component in the tracking by detection paradigm. 
Deep learning approaches for 2D object detection have improved drastically \cite{Lin2017FeaturePN, Ren_rrc2017, sun2018fishnet} since Faster R-CNN \cite{Ren_2017}. 
3D object detection receives increasing attention recently. 
To exploit both image and point cloud, some methods \cite{MV3D, AVOD} aggregate point cloud and image features from different views, while F-PointNet \cite{fpointnet} obtains frustum proposal from an image, and then applies PointNet \cite{Charles_2017} for 3D object localization with the point cloud. 
There exist state-of-the-art methods \cite{PointPillars, PointRCNN, VoxelNet} that use point cloud only. One-stage detectors \cite{PointPillars, VoxelNet} usually apply CNN on the voxelized representation, and two-stage detectors such as Point RCNN \cite{PointRCNN} generates proposals first by segmentation, which are refined in the second stage. Our mmMOT is readily adaptable for both 2D and 3D object detectors.

\begin{figure*}[htb]
\begin{center}
\includegraphics[width=\linewidth]{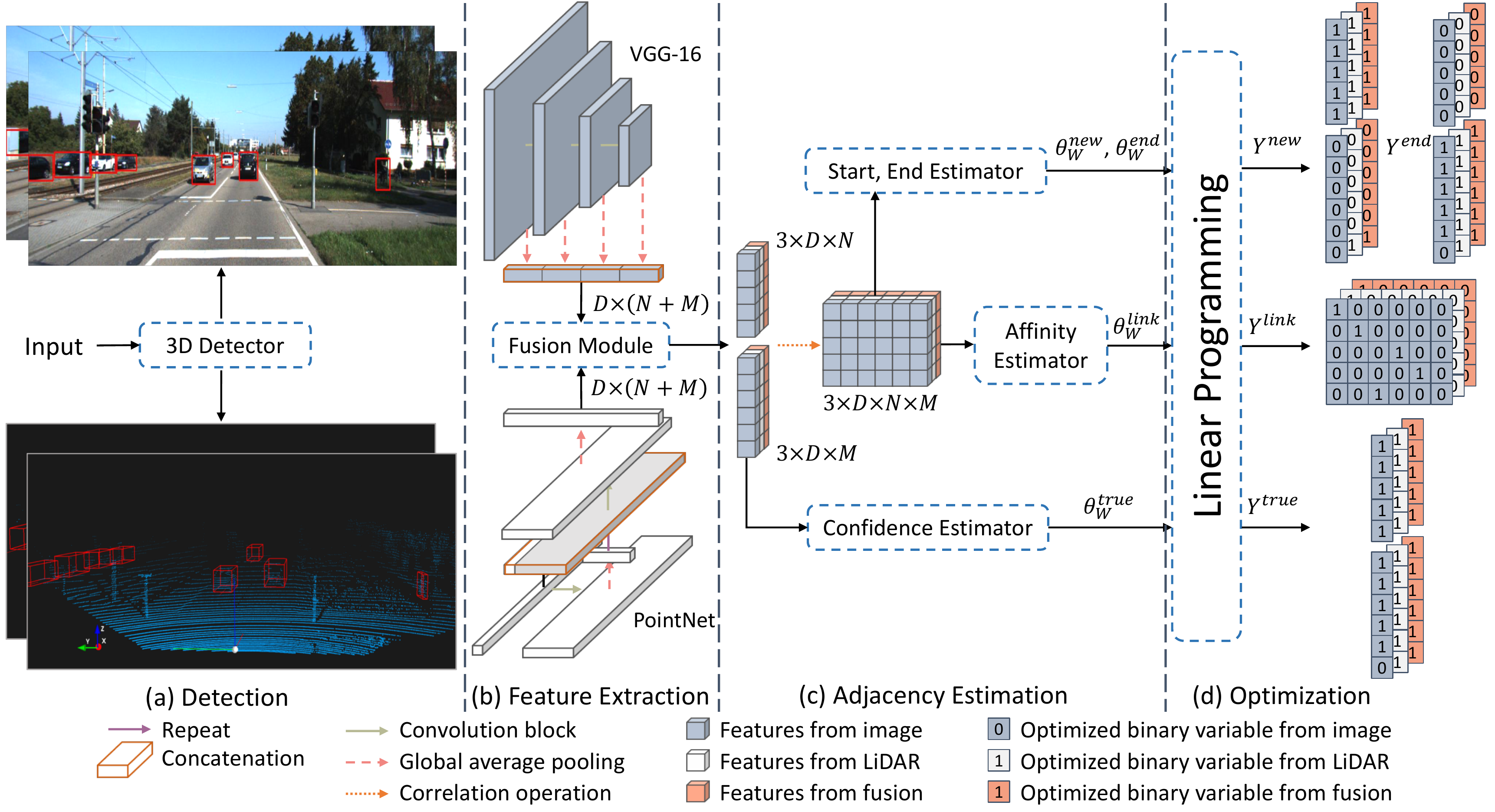}\\
 \vspace{-10pt}
 \end{center}
    \caption{The pipeline of mmMOT. The feature extractors first extract features from image and LiDAR, and the robust fusion module fuses the multi-sensor features. Next, the correlation operator produces the correlation features for each detection pair, by which the adjacency estimator predicts the adjacency matrix. All the predicted scores are optimized to predict the binary variable $Y$.}
 \vspace{-8pt}
 \label{fig:overall}
 \end{figure*}
 
\section{Multi-Modality Multi-Object Tracking}
 
We propose a multi-modality MOT (mmMOT) framework, which preserves reliability via independent multi-sensor feature extraction and improves accuracy via modality fusion. It generally follows the widely adopted tracking-by-detection paradigm from the min-cost flow perspective.
Specifically, our framework contains four modules including an object detector, feature extractor, adjacency estimator and min-cost flow optimizer, as shown in Fig. \ref{fig:overall} (a), (b), (c), (d), respectively. 
First, an arbitrary object detector is used to localize objects of interests. We use PointPillar \cite{PointPillars} for convenience. 
Second, the feature extractor extracts features from each sensor independently for each detection (Section \ref{sec:feature}), after which a fusion module is applied to fuse and pass the single modality feature to the adjacency estimator (Section \ref{sec:fusion}).  
The adjacency estimator is modality agnostic. It infers the scores necessary for the min-cost flow graph computation. The structure of the adjacency estimator and the associated end-to-end learning method will be demonstrated in Section \ref{sec:adjacent_matrix}.  
The min-cost flow optimizer is a linear programming solver that finds the optimal solution based on the predicted scores (Section \ref{sec:optim}). 

\subsection{Problem Formulation}
\label{sec:formulation}
Our mmMOT follows the tracking-by-detection paradigm to define the data association cost, which is solved as a min-cost flow problem \cite{FrossardU18, Lenz_2015, Schulter_2017}. 
Take the online MOT setting for example, assume there are $N$ and $M$ detections in two consecutive frames $i$ and $i+1$, denoted by $X^i= \left \{ x_{j}^{i}\mid j=1,\cdots, N \right \}$ and $X^{i+1}= \left \{ x_{k}^{i+1}\mid k=1,\cdots, M \right \}$, respectively. 
Each detection is associated to four types of binary variables in this paradigm. We introduce them following the notation of Deep Structured Model (DSM) \cite{FrossardU18}. 
First, for any $x_j$, a binary variable $y_j^{true}$ indicates whether the detection is a true positive. 
Second, binary variables $y_{jk}^{link}$ indicates if the $j$-th detection in the first frame and $k$-th detection in the second frame belong to the same trajectory, and all these  $y_{jk}^{link}$ form an adjacency matrix $A^i \in R^{N\times M}$, where $A_{jk}^i = y_{jk}^{link}$. 
The other two variables $y_j^{new}$, $y_j^{end}$ represents whether the detection is the start or the end of a trajectory, respectively. 
For convenience, we flatten the adjacency matrix into a vector $Y^{link}$, and gather all the binary variables having the same type as $Y^{true}$, $Y^{new}$, $Y^{end}$, then all these variables are collapsed into a vector 
$\mathbf{Y} = \left [Y^{true}, Y^{link}, Y^{new}, Y^{end} \right ]$, which comprises all states of edges in the network flow.
For each binary variable in $Y^{true}$, $Y^{link}$, $Y^{new}$, $Y^{end}$, the corresponding scores are predicted by the confidence estimator, affinity estimator, start and end estimator, respectively. These estimators form the adjacency estimator, and we solve them in a multi-task learning network as shown in Figure \ref{fig:overall}.

\begin{figure}
 \begin{center}
\includegraphics[width=\linewidth]{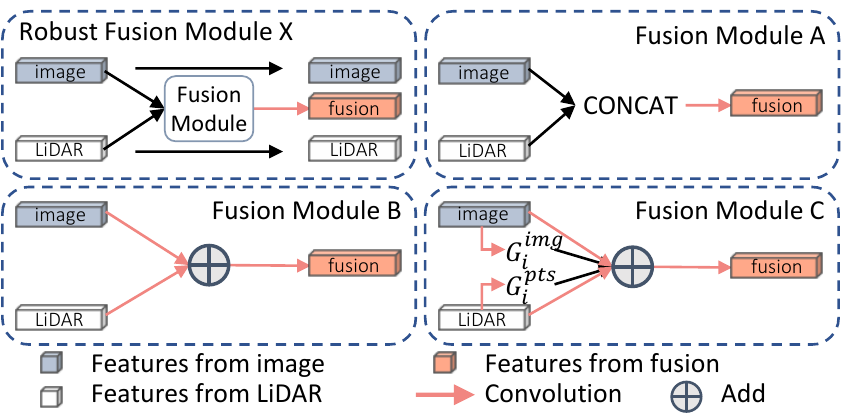} 
\vspace{-18pt}
 \end{center}
    \caption{The robust fusion module and three multi-modality fusion modules. The robust fusion module can apply any one of the fusion modules A, B and C to produce the fused modality. Unlike the conventional fusion modules, the robust fusion module produces both the single modalities and the fused modality as an output. Fusion module A concatenates the multi-modality features, module B fuses them with a linear combination, module C introduces attention mechanism to weights the importance of sensor's feature adaptively.}
 \label{fig:fusion}
 \vspace{-12pt}
 \end{figure} 

\subsection{Single Modality Feature Extractor}
\label{sec:feature}
In an online setting, only detections in two consecutive frames are involved. To estimate their adjacency, their deep representations are first extracted from the respective image or point cloud. 
The features of each single modality form a tensor with a size of $1 \times D \times \left( N+M\right )$, where $D=512$ is the vector length, and $N+M$ is the total number of detections in the two frames. 

\noindent\textbf{Image Feature Extractor}.
Upon obtaining 2D bounding boxes from either a 2D or 3D detector, the image patches associated to each detection are cropped and resized to a square with a side length of 224 pixels to form a batch. 
All these patches form a 4D tensor with a size of $\left( N+M\right ) \times 3 \times 224\times 224$.
We use VGG-Net \cite{SimonyanZ14a_vgg} as the image feature extractor's backbone.  
To exploit features at different level, we modify the skip-pooling \cite{Bell_2016} so as to pass different level's feature to the top, as shown in the VGG-Net depicted in Figure \ref{fig:overall}. 
The details of skip-pooling are provided in the supplementary material.

\noindent\textbf{Point Cloud Feature Extractor}. 
One of our contributions is to apply deep representation of point cloud in data association process for MOT.
While the LiDAR point cloud associated to a single detection could be easily obtained by a 3D bounding box, it remains a challenge if only a 2D bounding box is provided.
It is possible to obtain a 3D bounding box using F-PointNet \cite{fpointnet}, or directly estimated the 3D bounding box with other geometric information and priori \cite{Scheidegger_2018, BeyondPixels_ICRA2018}.
In this study, we choose not to locate the detection in 3D space because we observed more errors. Rather, inspired by F-PointNet \cite{fpointnet}, we exploit all the point clouds in the frustum projected by the 2D bounding box. This leads to high flexibility and reliability, and save computation from obtaining 3D bounding box.

The point cloud forms a tensor with a size $1\times D\times L$, where $L$ is the total number of all the points in all bounding boxes, and $D=3$ is the dimension of the point cloud information. We empirically found the reflectivity of point cloud provides only with marginal improvement, thus we only used the coordinates in 3D space.
We modify the vanilla PointNet \cite{qi2017pointnet} to extract features from point cloud for each detection as shown in the PointNet depicted in Figure \ref{fig:overall}. 
To enhance the global information of points in each bounding box, we employ the global feature branch originally designed for the segmentation task in PointNet \cite{qi2017pointnet}, and we found that average pooling works better than max pooling in PointNet for tracking.
During pooling, only the feature of points belonging to the same detection are pooled together. The feature vector of point cloud has a length of 512 for each detection.

\subsection{Robust Multi-Modality Fusion Module}
\label{sec:fusion}
In order to better exploit multi-sensor features while maintaining the ability to track with each single sensor, our robust fusion module is designed to have the capability of fusing features of multiple modalities as well as handing original features from just a single modality.
 
\noindent\textbf{Robust Fusion Module}. 
The operations in the adjacency estimator is batch-agnostic, thus we concatenate single modalities and the fused modality in the batch dimension to ensure that the adjacency estimator could still work as long as there is an input modality.
This design enables the proposed robust fusion module to skip the fusion process or fuse the remaining modalities (if there are still multiple sensors) during sensor malfunctioning, and pass them to the adjacency estimator thus the whole system could work with any sensor combination.
Formally, we denote the feature vectors of different modalities as $\left \{ F_i^{s}\right \}_{s=0}^{S}$, where the number of sensors is $S$, and the fused feature is denoted as $F_i^{fuse}$. 
In our formulation, the features of fused modality has the same size as a single modality. The robust fusion module concatenates $\left \{ F_i^{s}\right \}_{s=0}^{S}$ and $F_i^{fuse}$ along the batch dimension and feeds them to the adjacency estimator. They form a tensor of size $\left(S+1\right) \times D \times \left(N+M\right)$.

The robust fusion module could employ arbitrary fusion module, and we investigate three fusion modules as shown in Figure \ref{fig:fusion}. 
Take two sensors' setting as an example, the fusion module A naively concatenates features of multiple modalities; the module B add these features together; the module C introduces attention mechanism. 

\noindent\textbf{Fusion Module A}. A common approach is to concatenate these features, and use point-wise convolution with weight $W$ to adapt the length of the output vector to be the same as a single sensor's feature as follows:
\begin{equation}
    F_i^{fuse} = W \otimes \textrm{CONCAT}\left( F_i^{0}, \cdots ,F_i^{S} \right ),
\end{equation}
where $\otimes$ denotes a convolution operation, and $\textrm{CONCAT}\left(\cdot  \right )$ denotes a concatenation operation. 

\noindent\textbf{Fusion Module B}. Another intuitive approach is to fuse these two features with addition, we reproject the features of each modality and add them together as follows:
\begin{equation}
    F_i^{fuse} = \left( \sum\nolimits_{s=0}^{S} W^{s} \otimes F_i^{s} \right),
\end{equation}
where $W^{s}$ denotes the corresponding convolution kernels to the $s$-th sensor's feature. By addition the module gathers information from each sensor, and correlation feature of fused modality is also more like that of single sensor. It is favorable for the adjacency estimator to handle different modality since the correlation operation is multiplication or subtraction.

\noindent\textbf{Fusion Module C}. The module C introduces an attention mechanism for guiding the information fusion from different sensors, since the significance of a sensor's information might vary in different situations, e.g., the point cloud feature might be more important when the illumination condition is bad, and the image feature might be more important when the point cloud is affected in rainy days. 
The attention map $G_i^{s}$ for each sensor is first calculated as follows:
\begin{equation}
     G_i^{s} = \sigma \left ( W_{att}^{s} \otimes F_i^{s} \right ),
\end{equation}
where $W_{att}^{s}$ is the convolution parameter and $\sigma$ is a sigmoid function. 
We expect the $W_{att}^{s}$ to learn predict the importance conditioned on the feature itself, and the sigmoid function normalizes the attention map to be in the range from 0 to 1.
Then the information is fused as follows:
\begin{equation}
    F_i^{fuse} = \frac{1}{\sum_{s=0}^{S} G_i^{s}}\sum\nolimits_{s=0}^{S} G_i^{s} \odot \left( W^{s} \otimes F_i^{s} \right),
\end{equation}
where $\odot$ denotes element-wise multiplication, and the summation of $G_i^{s}$ is taken as a denominator for normalization.

\subsection{Deep Adjacency Matrix Learning}
\label{sec:adjacent_matrix}
Given the extracted multi-modality features, the adjacency estimator infers the confidence, affinity, start and end scores in the min-cost flow graph \cite{FrossardU18, Schulter_2017} based on each modality.
These features are shared for each branch of the adjacency estimator, namely the confidence estimator, affinity estimator, start and end estimator.
It is straightforward to learn a model for confidence estimator by taking it as a binary classification task. We focus on the design of the two other branches. 

\noindent\textbf{Correlation Operation}. To infer the adjacency, the correlation of each detection pair is needed. 
The correlation operation is batch-agnostic thus it can handle cross-modality, and the operation applied channel by channel to take advantage of the neural network. 
The commutative property is theoretically favorable for learning paired data, since it is  agnostic of the order of $ F_{j}^{i}$ and $F_{k}^{i+1}$. 
In this work, we compare three simple yet effective operators as follows:
\begin{packed_itemize}
    \item Element-wise multiplication,: $F_{jk}=F_{j}^{i} \odot  F_{k}^{i+1}$,
    \item Subtraction: $ F_{jk}=F_{j}^{i} - F_{k}^{i+1}$,
    \item Absolute subtraction: $ F_{jk}=\left | F_{j}^{i} - F_{k}^{i+1} \right | $.
\end{packed_itemize}
The element-wise multiplication is equivalent to a depthwise correlation filter \cite{li2018siamrpn}, where the filter size is $1\times1$.
The subtraction measures the distance of two vectors.
By taking the absolute value of subtraction, the operation becomes commutative and agnostic to the chronology of detection, which makes the network more robust. 

\begin{figure}
    \begin{center}
    \includegraphics[width=\linewidth]{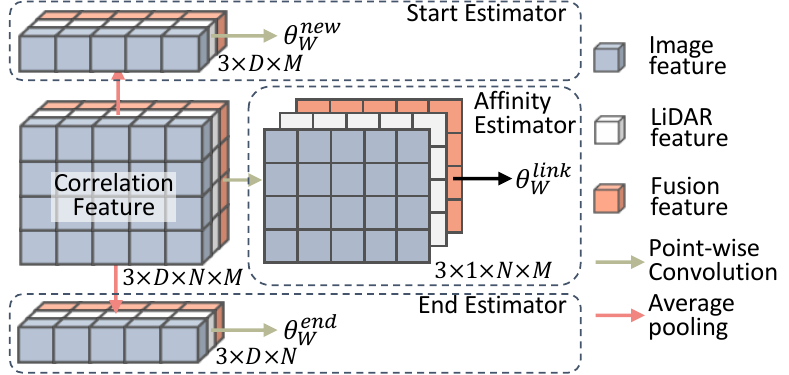}\\
     \vspace{-6pt}
    \end{center}
    \caption{The structure of the affinity estimator and start and end estimator. The affinity estimator estimates the adjacency using point-wise convolution. The start and end estimator gathers the correlation feature of each detection to check whether a detection is linked to make prediction more robust.}
    \label{fig:end_estimator}
     \vspace{-12pt}
\end{figure}

\noindent\textbf{Affinity Estimator}.
The obtained $F_{jk}$ is then used by the affinity estimator to predict the adjacency matrix $A^i$.
Since the correlation operation handles multi-modality in the batch dimension and is performed on each detection pair between two frames, the correlation feature map has a size of $3 \times D \times N \times M$.
We use 2D point-wise convolution as shown in Figure \ref{fig:end_estimator}. 
This makes the network handle each correlation feature separately since it only needs to determine whether $F_{jk}$ indicates a link. 
Since the convolution is batch-agnostic, it could work on any combination of modality, and the output adjacency matrix will has a size of $3 \times 1 \times N \times M$. 
Because these three predictions have the same target, we apply supervision signal to each of them, which enables joint optimization of feature extractor for each modality and affinity estimator for cross modality. 
During inference, the affinity estimator needs no modification if the sensor combination is changed, which allows both flexibility and reliability.

\noindent\textbf{Start and End Estimator}.
The start and end estimator estimate whether a detection is linked, thus their parameters are shared for efficiency.
Given the correlation feature $F_{jk}$, after gathering all the correlation information of one detection in each row or column by average pooling, the estimator also uses point-wise convolution to infer whether one detection is linked as shown in Figure \ref{fig:end_estimator}. 
Since the pooling layer is batch-agnostic, the start and end estimator is also flexible for different sensor settings. During inference, we simply pad zeros for new score of detection in the first frame and end score of detection in the last frame, since they cannot be estimated from the correlation feature map.

\noindent\textbf{Ranking Mechanism}.
We denote the raw output of the neural network's last layer as $o_{jk}^i$,
and we found that $a_{jk}$ should also be the greatest value among $a_{js}, s=1,...M$ and $a_{tk}, t=1,...N$, but directly take $A_{jk}^i=o_{jk}^i$ does not exploit this global information, thus we design a ranking mechanism to handle this problem.
Specifically, we apply a softmax function for each row and each column in the output matrix, and gather these two matrices to get the final adjacency matrix. 
In this work, we investigate three operations to combine the two softmax feature maps: max, multiplication and average. Taking the multiplication for example, the ranking mechanism is introduced as follows:
\begin{equation}
    a_{jk}^i= \frac{e^{o_{jk}^i}}{\sum_{s=0}^{N}e^{o_{js}^i}} \times \frac{e^{o_{jk}^i}}{\sum_{t=0}^{M}e^{o_{tk}^i}}.
\end{equation}

\noindent\textbf{Loss Function}.
The whole framework can be learnt in an end-to-end manner in a multi-task learning framework. We adopt the cross entropy loss for the classification branch and the L2 loss for the other two, thus the overall loss function can be written as follows: 
\begin{equation}
    L = L_{link} + \alpha L_{start} + \gamma L_{end} + \beta L_{true},
\end{equation}
where $\alpha$, $\gamma$ and $\beta$ indicates the weight of loss for each task. We empirically set  $\alpha = \gamma=0.4$ and $\beta=1.5$ in all the experiments in this paper.

\subsection{Linear Programming}
\label{sec:optim}
After obtaining the prediction score from the neural networks, the framework needs to find an optimal solution from the min-cost flow graph.
There are several facts that could be exploited as linear constraints among these binary variables in $\mathbf{Y}$. Firstly, if a detection is a true positive, it has to be either linked to another detection in the previous frame or the start of a new trajectory. Therefore, for one detection in current frame and all detections in its previous frame, a linear constraint can be defined in the form as follows:
\begin{equation}
    \forall k, y_{k}^{true} = \sum\nolimits_{j=0}^{N}y_{jk}^{link} + y_{k}^{start}.
\end{equation}
Symmetrically, for one detection in previous frame and all detections in current frame, a linear constraint can be defined as follows: 
\begin{equation}
    \forall j, y_{j}^{true} = \sum\nolimits_{k=0}^{M}y_{jk}^{link} + y_{j}^{end}.
\end{equation}
These two constraints can be collapsed in a matrix form to yield $\mathbf{C}\mathbf{Y}=0$, which has already encodes all valid trajectories.
Then the data association problem is formulated as an integer linear program as follows:
\begin{equation}
\begin{split}
    \mathop{\arg\max}_{y} = \mathbf{\Theta\left ( X \right )}^\top \mathbf{Y} \\
    \textrm{s.t. } \mathbf{C}\mathbf{Y}=0, \mathbf{Y} \in \left\{0, 1 \right \}^{\left| \mathbf{Y} \right|},
\end{split}
\end{equation}
where $\mathbf{\Theta\left (X\right )}$ is a flattened vector comprising all the predicted scores by the adjacency estimator.

\section{Experiments}
\noindent\textbf{Dataset.}
Our method is evaluated on the challenging KITTI Tracking Benchmark \cite{Geiger2012CVPR}. This dataset contains 21 training sequences and 29 test sequences. We select 10 sequences from the training partition as the training set and the remaining 11 sequences as the validation set. The train/validation set split is entirely based on frame number of these sequences to make the total frame number of training set (3975) close to that of validation set (3945). We submit our test-set result with the model trained only on split training set for fair comparison \cite{Scheidegger_2018}. 

Each vehicle in the dataset is annotated with 3D and 2D bounding boxes with a unique ID across different frames, and this allows us to obtain the ground truth adjacency matrix for each detection predicted by the detector. We calculate the Intersection over Union (IoU) between each detection and ground truth (GT) bounding boxes, and assign the ID of one GT box to a detection if one has an IoU greater than 0.5 and has the greatest IoU among other detections. This setting is consistent with the test setting of KITTI Benchmark.
The KITTI Benchmark \cite{Geiger2012CVPR} assesses the performance of tracking algorithms relying on standard MOT metrics, CLEAR MOT \cite{Bernardin_clearMOT} and
MT/PT/ML \cite{LiHN09}. 
This set of metrics measures recall and precision of detection, and counts the number of identity switches and fragmentation of trajectories. 
It also counts the mostly tracked or mostly lost objects, and provides an overall tracking accuracy (MOTA).

\noindent\textbf{Implementation Details.}
We first produce detections using the official code of PointPillar \footnote[1]{\url{https://github.com/nutonomy/second.pytorch}} \cite{PointPillars}.
The whole tracking framework is implemented with PyTorch \cite{paszke2017automatic}. 
The image appearance model's backbone is VGG-16 \cite{SimonyanZ14a_vgg} with Batch Normalization \cite{IoffeS15_bn} pretrained on ImageNet-1k \cite{ILSVRC15}. 
For linear programming, we use the mixed integer programming (MIP) solver provided by Google OR-Tools \footnote[2]{\url{https://developers.google.com/optimization}}.
We train the model for 40 epochs using ADAM optimizer with a learning rate of $6e^{-4}$ and the super convergence strategy \cite{smith2017superconvergence}. 
We manually set the score to be $-1$ if the confidence score falls below 0.2, this forces any detection having low confidence to be ignored during linear programming.

\begin{table}[t]\small
\caption{Comparison of different modalities. `Frustum' indicates using point cloud in the frustum. Robust Modules X indicates using fusion module X in the robust fusion module.}
\vspace{-6pt}
\label{tab:point_cloud_image}
\addtolength\tabcolsep{-0.3em}
\begin{center}
    \begin{tabular}{c|l|cccc}
    \hline
    Method&Modality&MOTA$\uparrow$&ID-s$\downarrow$ &FP$\downarrow$&FN$\downarrow$ \\
    \hline \multirow{4}{*}{Baseline}
    &Image          &74.88  &454    &951    &1387 \\
    &Frustum        &75.50  &387    &918    &1418 \\
    &Point Cloud    &75.70  &362    &946    &1393 \\
    &Ensemble       &77.54  &158    &949    &1388 \\
    \hline \multirow{3}{*}{Robust Module A}
    &Image          &75.40  &396    &951    &1387 \\
    &Point Cloud    &76.13  &317    &948    &1392 \\
    &Fusion         &77.57  &177    &910    &1406 \\
    \hline \multirow{3}{*}{Robust Module B}
    &Image          &75.17  &421    &951    &1387 \\
    &Point Cloud    &74.55  &490    &951    &1387 \\
    &Fusion         &77.62  &193    &\textbf{850}    &1444 \\
    \hline \multirow{3}{*}{Robust Module C}
    &Image          &74.86  &456    &951    &\textbf{1387} \\
    &Point Cloud    &74.94  &452    &946    &1398 \\
    &Fusion         &\textbf{78.18}  &\textbf{129}  &895    &1401 \\
    \hline Module A
    &Fusion         &77.31  &176    &934    &1412 \\
    Module B
    &Fusion         &77.31  &212    &913    &1396 \\
    Module C
    &Fusion         &77.62  &142    &945    &1400 \\
    \hline
\end{tabular}
\end{center}
\vspace{-18pt}
\end{table}

\subsection{Ablation Study}
To evaluate the proposed approach and demonstrate the effectiveness of the key components, we conduct an ablation study on the KITTI benchmark \cite{Geiger2012CVPR} under the online setting, with the state-of-the-art detector PointPillar \cite{PointPillars}.
We found that PointPillar detector produces large amount of false positive detections with low prediction score, so we discard detections with a score below 0.3. This does not hurt the mAP of detection, but saves a lot of memory in training.

\noindent\textbf{Competency of Point Cloud for Tracking}.
We set a 2D tracker as our baseline, which only employs 2D image patches as cues and use multiplication as correlation operator during data association, without the ranking mechanism.
We first compare the effectiveness of image and LiDAR point cloud, and evaluate two approaches to employ the point cloud: using point cloud in the frustum or in the bounding box.
From the row of baseline in Table \ref{tab:point_cloud_image}, it is observed that using the point cloud in the frustum yields competitive results as using that in the bounding box. The results suggest the applicability of point cloud even with 2D detections (as discussed in Section \ref{sec:feature}), thus the proposed framework is adaptable for 2D or 3D detector with arbitrary modality. 
More surprisingly, all point cloud methods perform better than the image baseline, which suggests the efficacy of point cloud's deep representation, and indicates that the system could still work when camera is failed.

\noindent\textbf{Robust Multi-Modality Fusion Module.}
We compare the effectiveness of the robust fusion modules A, B, and C. Baselines comprise of trackers using a single sensor, i.e, camera or LiDAR; we train and evaluate each modality separately.
To form a stronger baseline, we ensemble the image model (MOTA $74.88$) and point cloud model in bounding box  (MOTA $75.70$), and yields much better result (MOTA $77.54$). As shown in Table \ref{tab:point_cloud_image}, only robust fusion module C with attention mechanism surpasses the ensemble result remarkably, although all fusion methods surpass single-sensor baselines. The results suggest the non-triviality of finding a robust fusion module for multi-sensor inputs.

Since each methods with robust fusion module also provides prediction of single sensor, we compare the single sensor results of each robust fusion module in Table \ref{tab:point_cloud_image}. As can be observed, while the proposed Robust Module is capable of fusing multi-modality effectively, it can maintain competitive performance on the single modality in comparison to baselines (wherein dedicated training on single modality is conducted). Such kind of reliability in fusion is new in the literature.

\begin{table}[t]
\caption{Comparison of 2D trackers with further modification.}
\vspace{-6pt}
\label{tab:modification}
\begin{center}
\addtolength\tabcolsep{-0.2em}
    \begin{tabular}{l|cccc}
    \hline
    Modification &MOTA$\uparrow$&ID-s$\downarrow$ &FP$\downarrow$&FN$\downarrow$ \\
    \hline
    Multiplication      &74.88  &454    &951    &1387 \\
    Subtraction         &75.27  &410    &951    &1387 \\
    Absolute subtraction&\textbf{77.76}    &\textbf{143}    &941    &\textbf{1387} \\   
    \hline
    Softmax w mul       &75.08  &431    &951    &1387 \\
    Softmax w max       &76.24  &313    &940    &1387 \\
    Softmax w add       &77.40  &234    &\textbf{891}    &1387 \\
    \hline
\end{tabular}
\end{center}
 \vspace{-12pt}
\end{table}

\begin{table}[htbp]
\caption{Further improvement on fusion results. 'Correlation' indicates using absolute subtraction as correlation operation, 'Ranking' indicates using softmax with addition in ranking mechanism.}
\vspace{-6pt}
\label{tab:improve}
\begin{center}
\addtolength\tabcolsep{-0.2em}
    \begin{tabular}{cccccccc}
    \hline
    Correlation&Ranking&MOTA$\uparrow$&ID-s$\downarrow$ &FP$\downarrow$&FN$\downarrow$ \\
    \hline
                &           &78.18  &129    &895    &1401 \\
    \checkmark  &           &79.18  &23     &873    &1418 \\
    \checkmark  &\checkmark &\textbf{80.08}  &\textbf{13}     &\textbf{790}    &1411 \\
    \hline
\end{tabular}
\end{center}
\vspace{-18pt}
\end{table}

\begin{table*}[htbp]\small
\caption{Comparison on the testing set of KITTI tracking benchmark. Only published online methods are reported.}
\label{tab:sota}
\begin{center}
    \begin{tabular}{l|ccccccccccc}
    \hline
    Method &MOTA$\uparrow$&MOTP$\uparrow$&Prec.$\uparrow$&Recall$\uparrow$&FP$\downarrow$&FN$\downarrow$&ID-s$\downarrow$&Frag$\downarrow$&MT$\uparrow$&ML$\downarrow$\\
    \hline
    DSM \cite{FrossardU18} 
        &76.15&83.42&98.09&80.23&578&7328&296&868&60.00&8.31 \\
    extraCK \cite{GunduzA18}
        &79.99&82.46&98.04&84.51&642&5896&343&938&62.15&5.54 \\
    PMBM \cite{Scheidegger_2018} 
        &80.39&81.26&96.93&85.01&1007&5616&121&613&62.77&6.15 \\
    JCSTD \cite{tian2019online} 
        &80.57&81.81&98.72&83.37&405&6217&\textbf{61}&643&56.77&7.38 \\
    IMMDP \cite{XiangAS15}
        &83.04&82.74&\textbf{98.82}&86.11&\textbf{391}&5269&172&\textbf{365}&60.62&11.38 \\
    MOTBeyondPixels \cite{BeyondPixels_ICRA2018} 
        &84.24&\textbf{85.73}&97.95&88.80&705&4247&468&944&73.23&2.77 \\
    \hline
    mmMOT-normal &\textbf{84.77}&85.21&97.93&\textbf{88.81}&711&\textbf{4243}&284&753&\textbf{73.23}&\textbf{2.77} \\
    mmMOT-lose image&84.53&85.21&97.93&88.81&711&4243&368&832&73.23&2.77\\
    mmMOT-lose point cloud &84.59&85.21&97.93&88.81&711&4243&347&809&73.23&2.77\\
    \hline
\end{tabular}
\end{center}
 \vspace{-20pt}
\end{table*}

\begin{figure*}
\begin{center}
\includegraphics[width=\linewidth]{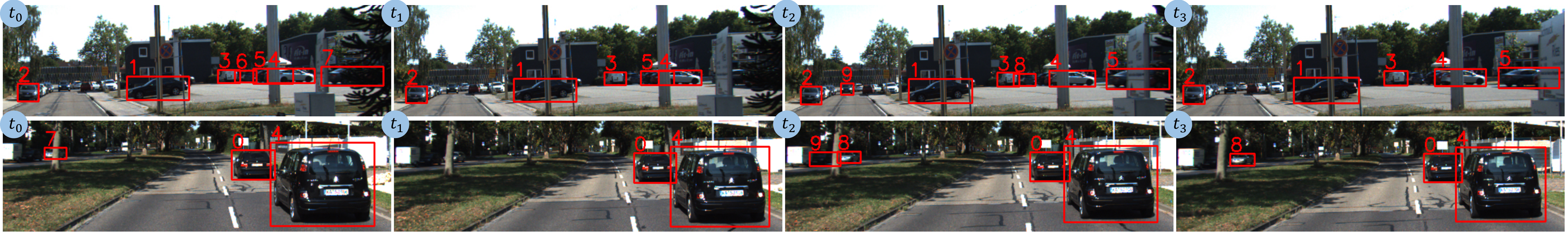} \\
 \vspace{-9pt}
 \end{center}
    \caption{Failure case analysis.}
 \label{fig:failure}
 \vspace{-14pt}
 \end{figure*}

\noindent\textbf{Fusion Module.}
We further compare the results of normal fusion modules, which only outputs fused modality to the adjacency estimator, thus the tracker cannot perform tracking with single modality under multi-modality setting. 
The results in the last row of Table \ref{tab:point_cloud_image} shows that the proposed robust module outperforms the baseline modules A, B, and C consistently, with the additional capability of handling single modality.
The results suggests that by preserving reliability, mmMOT gets more supervision signal which is favorable, and thus further improves the accuracy.

\subsection{Further Analysis}
\noindent\textbf{Correlation Operator.}
We further conduct experiments on correlation function discussed in Section \ref{sec:adjacent_matrix}, and compare the effectiveness of three different correlation functions on the 2D baseline. 
As shown in Table \ref{tab:modification}, the subtraction variant always performs better than the multiplication variant, and with commutative property the absolute subtraction performs the best.

\noindent\textbf{Ranking Mechanism}.
We also examine the effectiveness of the ranking mechanism, and investigate three different variants: the Softmax w mul, Softmax w max, Softmax w add, which indicate combining the softmax output by multiplication, argmax, addition, respectively. From Table \ref{tab:modification}, we can see that the ranking mechanism could improve MOTA by 0.2 at least, and adding the softmax output could yield improvement of about 2.5 in MOTA.

\noindent\textbf{Best Results with 3D Detection}.
We further improve the results of fusion model. Following the conclusion in Table \ref{tab:modification}, we use the absolute subtraction for correlation operation, and softmax activation by addition for ranking mechanism. We compare the efficacy of each modification in Table \ref{tab:improve}. The absolute subtraction correlation improves the fusion model's MOTA by 1, and the softmax activation with addition further improves 1 in MOTA and decreases the count of ID switches to 13, which is a remarkable improvement.

\subsection{KITTI Results}
We achieve state-of-the-art and competitive results using 2D detection from RRC-Net \cite{Ren_rrc2017} provided by MOTBeyondPixels \cite{BeyondPixels_ICRA2018}.
We use PointNet \cite{qi2017pointnet} to process point clouds in frustum, and VGG-16 \cite{SimonyanZ14a_vgg} for image patches. More details are provided in the supplementary material.
Table \ref{tab:sota} compares our method with other published state-of-the-art online methods. We first test mmMOT using all the modalities, namely mmMOT-normal. Then we simulate the sensor failure case by only passing single modality to the same model, named mmMOT-lose image/point cloud. Under both conditions our mmMOT surpass all the other published state-of-the-art online methods on MOTA.

The proposed method by modality fusion surpasses the previous best method MOTBeyondPixels \cite{BeyondPixels_ICRA2018} by far fewer ID switches (184 fewer) with the same detection method.
It is noteworthy that our single modality results still perform better, and we did not use bounding box and shape information of detections while MOTBeyondPixels does.
PMBM \cite{PosseggerMRB14}, JCSTD \cite{tian2019online}, and IMMDP \cite{XiangAS15} exhibit fewer ID switches but miss approximately one to two thousand detections. Those missed detections are hard examples not only for detection but also for tracking, so it is likely that they would exhibit higher number of ID switches than our method if they use the same detections.
Our method with each of the modalities surpasses the DSM \cite{FrossardU18} and extraCK \cite{GunduzA18} with fewer False Negatives and ID switches, i.e, our method makes fewer mistakes even when more hard examples are given.

\subsection{Failure Case Analysis}

We observe several conditions that could cause failure in our mmMOT. The statistical results are provided in the supplementary material, and the examples are shown in Figure \ref{fig:failure}, where each row includes four consecutive frames in a video.
First, for objects far away, early errors caused by 2D detector will lead to false negative detection as shown by the car with ID 9 in the first row. The error could also cause ID switches if the car is missed but recovered, as shown by the car with ID 6 in the first row and the car with ID 7 in the second row.
Second, the illumination also affects the performance, e.g., the black car in the shade with ID 9 in the second row.
Third, the occlusion also causes difficulties, e.g., the detector missed the car with ID 7 in the first row. 
And partial observation makes the cars hard to be distinguished, e.g., the cars with ID 5 and 7 in the first row both only have black rears observed, thus are inferred to be the same.
To further address the challenge caused by the occlusion, illumination and long distance, one may further exploit multi-modality in detection to prevent early errors, or exploit more information (e.g., temporal information) in data association to reinforce the prediction.

\section{Conclusion}
We have presented the \textbf{mmMOT}: a \textbf{m}ulti-\textbf{m}odality \textbf{M}ulti-\textbf{o}bject \textbf{T}racking framework. 
We make the first attempt to avoid single-sensor instability while keeping multi-modality effective via a deep end-to-end network. Such a function is crucial for safe autonomous driving and has been overlooked by the community.
Our framework is learned in an end-to-end manner with adjacency matrix learning, thus could learn to infer from arbitrary modality well in the same time.
In addition, this framework is the first to introduce deep representation of LiDAR point cloud into data association problem, and enhances the multi stream framework's robustness against sensor malfunctioning.

\noindent\textbf{Acknowledgment}
This work is supported by SenseTime Group Limited, Singapore MOE AcRF Tier 1 (M4012082.020), NTU SUG, and NTU NAP.

\setcounter{section}{0}
\setcounter{figure}{0}
\renewcommand{\thesection}{A\arabic{section}}
\renewcommand{\thetable}{A\arabic{table}}
\renewcommand{\thefigure}{A\arabic{figure}}
\section{Model Details}
\subsection{Skip Pooling}
The skip pooling layer passes the output feature of each max pooling layer (except the first) in the VGG-Net to the top. 
Specifically, the numbers of channels for the output of the max pooling layer are 64, 128, 256, 512 in VGG-Net. 
The global average pooling is first applied to these outputs to gather the spatial information in each level's feature. Next, we use two point-wise convolutions with normalization and ReLU activation to re-scale their number of channels to 128.
Then we concatenate these four vectors into a vector with length 512, which is taken as the image feature of each detected bounding-box in the following pipeline.

\subsection{Best Model}
\label{sec:mmmot_detail}
In the best model on KITTI test set, we still use VGG-16 \cite{SimonyanZ14a_vgg} with Batch Normalization \cite{IoffeS15_bn} as our image feature extractor's backbone, pretrained on ImageNet-1k \cite{ILSVRC15} by Pytorch  \cite{paszke2017automatic}.
The hyper-parameters of PointNet \cite{qi2017pointnet} are kept to be the same as before. 
For detection we use a 2D detector RRC-Net \cite{Ren_rrc2017}, which has higher recall and precision than the 3D detector PointPillar \cite{PointPillars}. Thus, we use the point cloud in the frustum for each detection. 
We use fusion module C to exploit image and point cloud stream, and use absolute subtraction as correlation operation, for ranking mechanism we use softmax activation with addition. 

\section{Failure Analysis}
\begin{figure}[ht]
\begin{center}
\includegraphics[width=\linewidth]{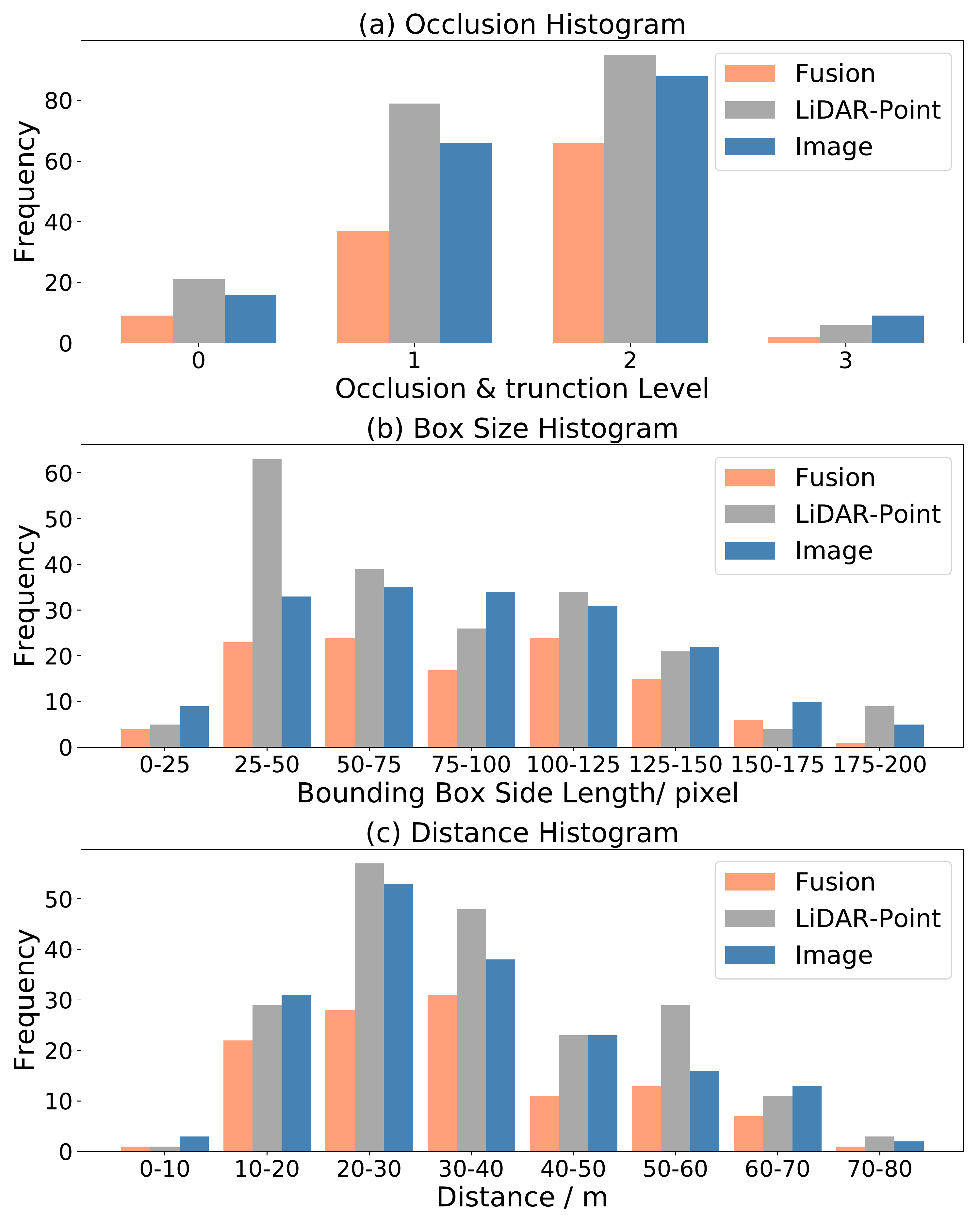} \\
 \end{center}
 \vspace{-6pt}
    \caption{Failure case analysis. Occlusion level 0, 1, 2, 3 indicates the object is not, moderately, highly, extremely occluded and truncated in image.}
 \label{fig:failure_statistic}
 \vspace{-12pt}
 \end{figure}

We further analyse the failure cases of our best mmMOT model with different modality. 
We focus on the amount of ID switches in the data association process, since the false negative and false positive are mainly caused by the detector. 
We analyse the occlusion condition, distance from ego car and the bounding box size of each object whose ID is switched.
The statistical results are shown in the Figure \ref{fig:failure_statistic}.
 
From Figure \ref{fig:failure_statistic} we observe that the fusion modality indeed makes the tracker more robust to more difficult occlusion and distance conditions.  
More interestingly, from Figure (a) we can observe that most of id switches come with occlusion, because partial observation could make the object hard to recognize or distinguish. 
And the occlusion causes more errors when only using point cloud than using image, because we use point cloud in the frustum for 2D detector, and more occlusion could also cause more points of occlusion in the frustum, which provide more information noise. 
From Figures (b) and (c) we observe that more errors come with small bounding box size and long distance, under which condition the objects' image patches are small and the point cloud is sparse. We also observe that point cloud modality faces more errors, because the number of points in small bounding box or at long distance is insufficient to represent the object, while the image patches could still be interpolated to have the size of $224 \times 224$.

{\small
\bibliographystyle{ieee_fullname}
\bibliography{mainbib}
}

\end{document}